# HUGS: Combining Exact Inference and Gibbs Sampling in Junction Trees

**Uffe Kjærulff**
Department of Mathematics and Computer Science, Aalborg University
Fredrik Bajers Vej 7E, DK-9220 Aalborg Ø, Denmark
uk@iesd.auc.dk

## Abstract

Dawid, Kjærulff & Lauritzen (1994) provided a preliminary description of a hybrid between Monte-Carlo sampling methods and exact local computations in junction trees. Utilizing the strengths of both methods, such hybrid inference methods has the potential of expanding the class of problems which can be solved under bounded resources as well as solving problems which otherwise resist exact solutions. The paper provides a detailed description of a particular instance of such a hybrid scheme; namely, combination of exact inference and Gibbs sampling in discrete Bayesian networks. We argue that this combination calls for an extension of the usual message passing scheme of ordinary junction trees.

## 1 INTRODUCTION

This paper presents an extension of the expert system shell Hugin (Andersen, Olesen, Jensen & Jensen 1989), called HUGS ("Hugin + Gibbs sampling"), involving a subset of the functionality described by Dawid et al. (1994). The extension involves the introduction of a new kind of belief universe, called a GIBBS universe. This has a dramatic impact on both the compilation process and the various inference steps performed in a junction tree.

Compiling a Bayesian network involves construction of a junction tree (Jensen 1988) via the processes of moralization and triangulation (Lauritzen & Spiegelhalter 1988). Inference in the Bayesian network is then formulated in terms of message passing in the junction tree (Jensen, Lauritzen & Olesen 1990). A node, $U$, of a junction tree is called a *belief universe* (or simply a *universe*) and consists of

- a cluster of variables of the Bayesian network and

- a set of potential functions (or simply potentials), $\Phi_U$, defined on these variables.

Provided the independence graph of the Bayesian network is a directed acyclic graph, $\Phi_U$ is initially a set of (conditional) probability functions, where each function is defined on a variable (the child) and a set of conditioning variables (the parents) such that both the child and its parents are members of $U$. Later, when messages are being passed to $U$ from its neighbours $V, \ldots, W$, $\Phi_U$ is extended with functions defined on $U \cap V, \ldots, U \cap W$, where we use $U$ etc. as shorthand for 'the variables associated with $U$'. These intersections are called *separators* (Jensen et al. 1990) (consult this reference for further details on the definition and properties of junction trees).

Hence, a universe, $U$, may be considered an autonomous entity containing local knowledge (i.e., $\Phi_U$). Absorption of messages possibly improves this knowledge, and eventually, when $U$ has received active messages from each of its neighbours, it has got sufficient information to calculate the true joint (probability) distribution over its variables. (A message is called *active* if the sender has received active messages from each of its neighbours, with the possible exception of the recipient, and it is the first message from the sender to the recipient with that property (Dawid 1992). Further, when an active message has been sent in each direction along each link of the junction tree, equilibrium has been established (i.e., each universe contains complete information to calculate the true joint distribution over its variables). The process of establishing equilibrium is also termed *propagation* and involves an inward and an outward pass with the inward pass being completed when only one universe is capable of sending an active message (in which case exactly one active message has been passed along each link of the junction tree).)

At any time, the exact joint potential, $\phi_U$, of a universe, $U$, is calculated as the product of the current potentials in $\Phi_U$. If all variables of $U$ are discrete and $U$ calculates $\phi_U$, we shall refer to $U$ as a DE universe, where 'DE' stands for 'discrete exact'. Note that the complexity of calculating $\phi_U$ may be expressed through the complexity of the complete subgraph induced by $U$ (or, in other words, the independence graph induced by $\phi_U$ is complete, since $\phi_U$ itself does not carry explicit information about its original factorization). In many real-world applications, the calculation of $\phi_U$ is prohibitive due to the size of $\mathcal{X}_U$ (the configuration space of $U$). Then a possible fruitful alternative (as suggested in the present paper) could be to approximate $\phi_U$ through simulation (i.e., establishing a potential $\hat{\phi}_U \approx \phi_U$ through repeated sampling of the potentials in $\Phi_U$).

If the calculation of the joint potential is approximated through Gibbs sampling (Geman & Geman 1984), we shall refer to $U$ as a GIBBS universe. Note that the complexity of calculating $\hat{\phi}_U$ is proportional to the complexity of the moral graph induced by $\Phi_U$ (i.e., the moral graph of the union of the independence graphs induced by the potentials in $\Phi_U$) provided that variables are only sampled simultaneously when necessary. However, to ensure convergence, it is often necessary to sample 'blocks' of variables simultaneously; thus in HUGS we use an advanced Gibbs sampling scheme (Jensen, Kong & Kjærulff 1995). We shall elaborate on the issue of blocking-Gibbs sampling in Section 2.

A message sent via a separator $S$ may be perceived as a list of configurations and associated weights. Configurations with no explicit weight associated are assumed to have weight zero. If the message contains a non-negative weight for each $x \in \mathcal{X}_S$, the message is *complete*; otherwise, it is *incomplete*. Note that in order to avoid conflicts, two or more incomplete messages cannot, in general, be absorbed simultaneously. This calls for an extension of the message-scheduling vocabulary and management as discussed in Sections 5 and 6.

The motivation for introducing GIBBS universes is twofold. First, as already discussed, the computational complexity imposed by a GIBBS universe is not exponential in the number of variables of the universe. Second, the use of Monte-Carlo type algorithms for inference makes it possible to handle a larger range of distributions and mixtures of distributions. In HUGS, however, we have limited the functionality to ordinary discrete-type Bayesian networks.

Thomas, Spiegelhalter & Gilks (1992) have described a general program, called BUGS, for Bayesian inference in graphical structures using Gibbs sampling. However, contrary to HUGS, BUGS relies exclusively on Gibbs sampling, but is able to handle a wide range of continuous as well as discrete distributions.

Section 3 presents a (small) example used throughout the paper to illustrate some relevant issues of propagation in HUGS. In Sections 4–8 we shall provide detailed descriptions of message format, message generation in a GIBBS universe, and message scheduling. Section 9 briefly discusses possible future extensions of HUGS.

## 2   GIBBS SAMPLING

When a GIBBS universe, $U$, has completed sampling (i.e., an approximation $\hat{\phi}_U$ of $\phi_U$ has been found), $U$ becomes a DE universe in the sense that all subsequent absorptions and generations of messages will be performed in the usual DE manner using $\hat{\phi}_U$ (or modifications of it). To see why $U$ cannot perform sampling in both the inward and the outward pass of propagation, it might be helpful to express $\hat{\phi}_U$ as the product of $\phi_U$ and a 'disagreement' function $\delta_U$ which can be considered an evidence (or likelihood) function. That is, in a sense, the error imposed when generating $\hat{\phi}_U$ amounts to inserting evidence into $U$. Then, obviously, performing sampling in the outward pass would require another propagation, which in turn would require yet another, etc. So, sticking to a two-pass approach, GIBBS universes must change status to DE whenever they have performed sampling.

Gibbs sampling is characterized by imposing a neighbouring structure onto the configuration space, say $\mathcal{X}_U$, from which the samples are drawn. That is, the samples are dependent in the sense that, given a particular current configuration, only a (small) subset of $\mathcal{X}_U$ will be candidate next configurations. The topology of the neighbouring structure depends heavily on the extent to which variables are sampled simultaneously: the greater the number of variables sampled simultaneously the denser the neighbouring structure. Further, the denser the neighbouring structure the larger the independence between samples and, consequently, the faster the convergence. So, finding an optimal balance between complexity and rate of convergence is essential.

Consider, for example, a universe, $U$, containing two binary variables $X_A$ and $X_B$ with $\mathcal{X}_A = \{a_1, a_2\}$ and $\mathcal{X}_B = \{b_1, b_2\}$. Assume that $\Phi_U = \{\varphi_U, \psi_B\}$, where $\varphi_U(a_1, b_1) = \varphi_U(a_2, b_2) = 1$, $\varphi_U(a_1, b_2) = \varphi_U(a_2, b_1) = 0$, and $\psi_B(X_B) = (z, 1 - z)$. Let $(a_1, b_1)$ be the initial configuration. Then sampling $X_A$ and $X_B$ individually, we get $\hat{\phi}_U(a_1, b_1) = 1$ and $\hat{\phi}_U(y) = 0$ for $y \in \mathcal{X}_U \setminus \{(a_1, b_1)\}$. Sampling $X_A$ and $X_B$ simultaneously, on the other hand, we get the correct answer, namely $\hat{\phi}_U(a_1, b_1) \to z$, $\hat{\phi}_U(a_2, b_2) \to 1 - z$, and $\hat{\phi}_U(a_1, b_2) = \hat{\phi}_U(a_2, b_1) = 0$. In the first case, the neighbouring structure is characterized by a disconnected graph, whereas, in the second case, it is a complete graph. The Markov chains induced are also said to be, respectively, reducible and irreducible.

Blocking-Gibbs sampling cannot, in general, guarantee irreducibility except, of course, in the degenerate case where all variables are sampled simultaneously. Various stochastic relaxation techniques, like simulated annealing, can be employed to guarantee irreducibility with probability arbitrarily close to 1, but this out of the scope of the present paper.

## 3 AN EXAMPLE: AUNT EMILY

The example chosen to illustrate the features of HUGS is the Aunt Emily network displayed in Figure 1 (a toy example for revealing the cause of death for two elderly, wealthy ladies who have heart troubles and unscrupulous and greedy heirs).

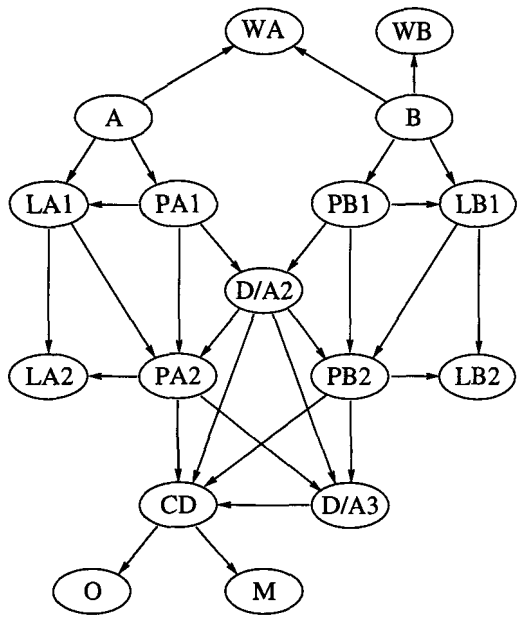

Figure 1: The Aunt Emily network.

A hybrid junction tree (i.e., a junction tree containing both DE and GIBBS universes) for the Aunt Emily network is displayed in Figure 2.

The distribution of potentials (conditional probability functions) over the various universes is handled as usual, except that the potentials are stored in different ways in the two kinds of universes and that GIBBS universes are preferred to DE universes for the sake of efficiency: the more information available before sampling in a GIBBS universe takes place, the fewer samples are required to obtain a certain level of precision.

For each DE universe, $U$, we calculate $\phi_U = \prod_{\varphi \in \Phi_U} \varphi$, where $\varphi : \mathcal{X}_V \to [0; \infty[$, $V \subseteq U$ and each $\varphi$ is extended to $\mathcal{X}_U$ (i.e., if $x \in \mathcal{X}_U$ and $y$ is the projection of $x$ on $\mathcal{X}_V$, then $\varphi(x) = \varphi(y)$). For each GIBBS universe, $U$, we simply store $\Phi_U$ in $U$. The distribution of potentials could be done as indicated in Table 1.

Table 1: Sample distribution of potentials showing in what form (i.e., product, $\phi_U$, or list, $\Phi_U$) the potentials are stored in the various universes.

| Universe | Potentials assigned |
|---|---|
| $DE_1$ | $\phi(LA1 \mid A, PA1) * \phi(PA1 \mid A)$ |
| $DE_2$ | $\phi(LA2 \mid LA1, PA2)$ |
| $DE_3$ | $\phi(LB1 \mid B, PB1)$ |
| $DE_4$ | $\phi(LB2 \mid PB2, LB1)$ |
| $DE_5$ | $\phi(M \mid CD)$ |
| $DE_6$ | $\phi(O \mid CD)$ |
| $GIBBS_1$ | $\{\, \phi(PB1 \mid B),\ \phi(D/A2 \mid PA1, PB1),$ $\phi(PA2 \mid LA1, PA1, D/A2)\,\}$ |
| $GIBBS_2$ | $\{\, \phi(A),\ \phi(B),\ \phi(WA \mid A, B),\ \phi(WB \mid B)\,\}$ |
| $GIBBS_3$ | $\{\, \phi(PB2 \mid D/A2, PB1, LB1),$ $\phi(D/A3 \mid PA2, D/A2, PB2),$ $\phi(CD \mid PA2, D/A2, D/A3, PB2)\,\}$ |

In conventional DE-type junction trees, propagation is typically performed at this point. The resulting universe potentials, which are true marginal distributions (provided normalization is performed in the 'root' universe when the inward pass has been completed), are used as the basis for subsequent belief revision.

In a hybrid junction tree, however, this initial propagation should not be performed, since the approximate DE representations of the potentials of the GIBBS universes will typically be crude approximations of the exact potentials. Therefore, using these approximate potentials as the basis for subsequent belief revision might lead to severely distorted posteriors. In particular, the distributional tails are likely to be severely distorted or even non-existing; thus, if these tails play a crucial role in the propagation of subsequent evidence, the computed posteriors might be extremely unreliable.

Also, if new evidence is to be incorporated after propagation has been performed, the computed posteriors should be discarded before the new evidence is entered and propagated.

To discuss and shed some light on the issues related to message generation and message passing in a hybrid junction tree we shall describe a propagation scenario based on the junction tree of Figure 2 and with evidence LA1 = 1, D/A3 = no, and O = arsenic. Since we shall restrict ourselves to sending active messages, we must start sending from a leaf universe. Assume that universe GIBBS$_2$ sends the first message.

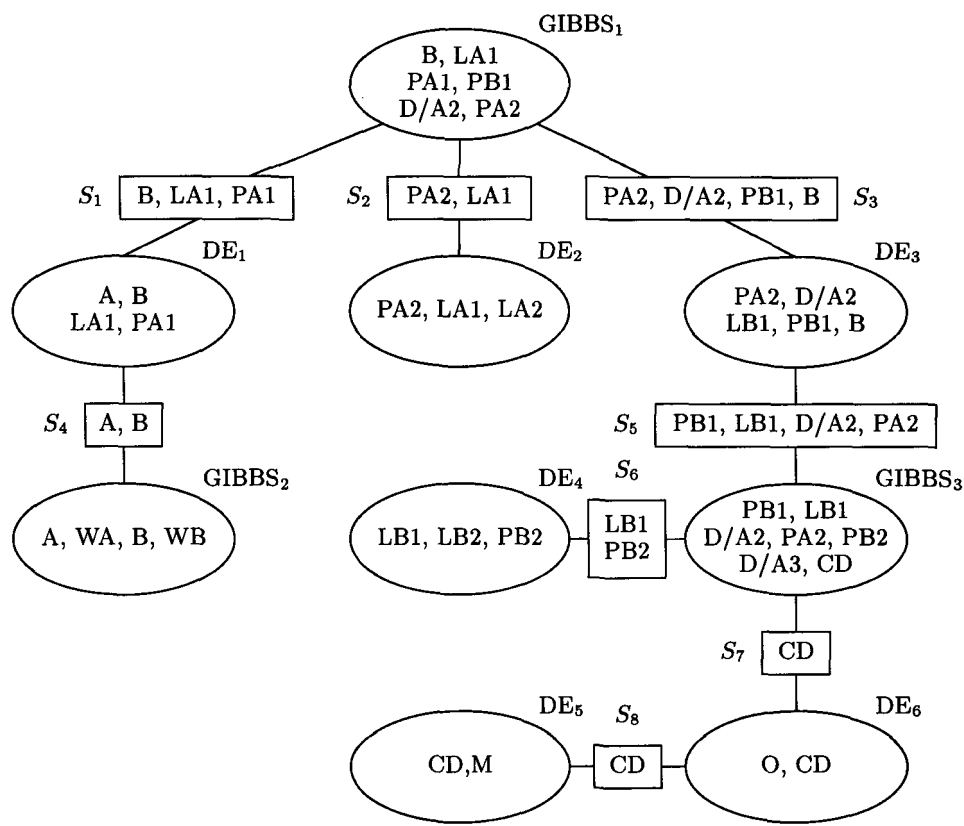

Figure 2: Hybrid junction tree for the Aunt Emily network.

## 4   MESSAGE FROM GIBBS$_2$

As discussed in Section 2, to send a message from a GIBBS universe, $U$, we must first turn it into a DE universe through sampling from the potentials in $\Phi_U$, producing an approximation $\hat{\phi}_U$ of $\phi_U$. Since the samples are generated using Gibbs-type sampling methods, the samples are obtained from a graphical structure (independence graph) based on the moral graph induced by the potentials in $\Phi_U$. (In case of plain Gibbs sampling (i.e., sampling one variable at a time), the moral graph is used; otherwise, performing blocking-Gibbs sampling, triangulated subgraphs of the moral graph are used (Jensen et al. 1995).)

### 4.1   CREATING SAMPLES

For simplicity, let us assume that plain Gibbs sampling is applied in our example. Thus, the relevant graphical structure is the subgraph of the directed acyclic graph in Figure 1 induced by A, B, WA, and WB. (Note

that in this case it is possible to sample from the true marginal distribution (cf. Table 1).)

Before being able to start sampling we must identify a legal initial configuration, say $x^0_{\text{GIBBS}_2}$, such that $\phi_{\text{GIBBS}_2}(x^0_{\text{GIBBS}_2}) > 0$. Ideally, $x^0_{\text{GIBBS}_2}$ should be found by drawing from $\phi_{\text{GIBBS}_2}$, but this might not be computationally tractable. Therefore, we shall limit ourselves to fulfilling the positivity requirement, implying that the first, say, 5–10% of the samples should be rejected such that sampling from $\phi_{\text{GIBBS}_2}$ can be expected; in the literature this rejection phase is often termed 'burn-in'. (Note that in this particular case an ideal initial configuration can easily be found through forward sampling, eliminating the need for burn-in.)

A legal initial configuration, $x^0_{\text{GIBBS}_2}$, can be searched for using either a deterministic or a stochastic approach. Using a deterministic approach, $x^0_{\text{GIBBS}_2}$ can be found by establishing the total orderings $\mathcal{X}_A = (x^{(0)}_A, x^{(1)}_A, x^{(2)}_A)$, $\mathcal{X}_B = (x^{(0)}_B, x^{(1)}_B, x^{(2)}_B)$, $\mathcal{X}_{WA} = (x^{(0)}_{WA}, x^{(1)}_{WA})$, $\mathcal{X}_{WB} = (x^{(0)}_{WB}, x^{(1)}_{WB})$, GIBBS$_2 = $ (A, B, WA, WB) and clamping the four variables to their (0)-state. Then, if all potentials in $\Phi_{\text{GIBBS}_2}$ are positive, we let $x^0_{\text{GIBBS}_2} = (x^{(0)}_A, x^{(0)}_B, x^{(0)}_{WA}, x^{(0)}_{WB})$; otherwise, clamp WB to $x^{(1)}_{WB}$. If the potentials are pos-itive, let $x^0_{\text{GIBBS}_2} = (x^{(0)}_A, x^{(0)}_B, x^{(0)}_{WA}, x^{(1)}_{WB})$; otherwise, clamp WA to $x^{(1)}_{WA}$ and WB to $x^{(0)}_{WB}$. Etc.

Having determined an $x^0_{\text{GIBBS}_2}$, the Gibbs sampler begins its job. Let the sampling order be, for example, A, B, WA, WB. That is, we need to compute

$$\phi(X_A \,|\, x^0_B, x^0_{WA}) \propto \phi(X_A) * \phi(x^0_{WA} \,|\, X_A, x^0_B)$$

and

$$\phi(X_B \,|\, x^1_A, x^0_{WA}, x^0_{WB}) \propto$$
$$\phi(X_B) * \phi(x^0_{WA} \,|\, x^1_A, X_B) * \phi(x^0_{WB} \,|\, X_B),$$

where $x^1_{\text{GIBBS}_2} = (x^1_A, x^1_B, x^1_{WA}, x^1_{WB})$ is the first sample produced. Note that $\phi(X_{WA} \,|\, x^1_A, x^1_B)$ and $\phi(X_{WB} \,|\, x^1_B)$ are readily available once A and B have been sampled.

## 4.2   WEIGHTS AND MESSAGE FORMAT

When the Gibbs sampler has completed its job and generated, say, $N$ samples, GIBBS$_2$ changes status from being a GIBBS universe to becoming a DE universe with the joint probability table $\hat{\phi}_{\text{GIBBS}_2}$ given by a set of $N' \leq N$ possible configurations $(x^1_{\text{GIBBS}_2}, \ldots, x^{N'}_{\text{GIBBS}_2})$ with weights $(w^1_{\text{GIBBS}_2}, \ldots, w^{N'}_{\text{GIBBS}_2})$, respectively, where $w^i_{\text{GIBBS}_2}$ equals the number of samples identical to $x^i_{\text{GIBBS}_2}$. Thus, the message, $\phi^*_{S_4}$, to be sent to DE$_1$ is of the form

$$\left( (x^1_{S_4}, w^1_{S_4}), \ldots, (x^{N''}_{S_4}, w^{N''}_{S_4}) \right), \qquad (1)$$

where $N'' \leq N'$ is the number of possible configurations in $\mathcal{X}_{S_4}$, and $\sum_i w^i_{S_4} = \sum_j w^j_{\text{GIBBS}_2} = N$. Ideally, the sum of weights should be $\sum \phi_{\text{GIBBS}_2}$ instead of $N$. However, since the calculation of this sum might be very time consuming, we shall attach weight 1 to each sample and perform appropriate normalisation afterwards (e.g. in the 'root' universe when the inward pass has been completed). As a consequence of this weighting scheme, the normalisation constant will no longer be useful.

## 5   CASCADING

The generation of the message $\phi^*_{S_4}$ by GIBBS$_2$ potentially reduces the set of possible configurations that other GIBBS universes are allowed to sample. This may happen only if $\phi^*_{S_4}$ is incomplete (i.e., $N'' < |\mathcal{X}_{S_4}|$) and only for GIBBS universes which either share variables with GIBBS$_2$ or contain variables related functionally to variables of GIBBS$_2$.

For example, imagine that we use a message scheduling in our sample junction tree where universe DE$_1$ become

'root' (i.e., DE$_1$ is the only universe capable of sending an active message when the inward pass has been completed) and that the message, $\phi^*_{S_1}$, from GIBBS$_1$ to DE$_1$ contains the information 'variable B cannot be in state $x^{(0)}_B$' (i.e., $\phi^*_{S_1}$ does not include configurations $x_{S_1}$ with $(x_{S_1})_B = x^{(0)}_B$). Then, if the message, $\phi^*_{S_4}$, from GIBBS$_2$ to DE$_1$ contains the information 'variable B can only be in state $x^{(0)}_B$' (i.e., $\phi^*_{S_4}$ does not include configurations $x_{S_4}$ with $(x_{S_4})_B \neq x^{(0)}_B$), the normalization constant in DE$_1$ becomes zero (cf. inconsistent evidence).

Therefore, in order to avoid such inconsistent messages, other GIBBS universes should possibly be notified of the constraints generated by a GIBBS universe before they start generating their own messages. The process of making them aware of such constraints is termed *cascading*, since we might need to initiate a cascade of messages to the relevant GIBBS universes. Note that cascading can involve non-active messages.

The determination of whether cascading is required or not can be quite complicated. Here we shall describe a simple yet sufficient, but not necessarily necessary cascading mechanism based on the observation that cascading will be required only if an incomplete message, $\phi^*_S$, is passed from one universe, $U$, to another universe, $V$, and the support of $\phi^*_S$ is smaller than the support of $\sum_{V \setminus U} \phi_V$ (i.e., there exists at least one configuration, $x_S$, such that $\phi^*_S(x_S) = 0$ and $\phi_V(x_S) > 0$). Note that incomplete messages may be generated by both GIBBS and DE universes. GIBBS universes (practically) always generate incomplete messages, whereas DE universes may only do so when they have received an incomplete message.

To be precise, having performed absorption (and possibly sampling) in a universe $U$, for each neighbour $V$ for which the message $\phi^*_{U \cap V}$ is incomplete, $\phi^*_{U \cap V}$ must be sent to $V$ as a cascade message unless

(1) the subtree rooted at $V$ (and not including the $U$-branch) does not contain a GIBBS universe,

(2) $U$ has already received a message from $V$, or

(3) all zeros of $\phi^*_{U \cap V}$ coincide with zeros of $\sum_{V \setminus U} \phi_V$.

Further, whenever a universe receives a cascade message, it must absorb the message (and possibly perform sampling), and all other message sending activity must be suspended (at least in a subtree) until the cascading has been completed.

Therefore, the message $\phi^*_{S_4}$ from GIBBS$_2$ to DE$_1$ is either an 'active non-cascade' or an 'active cascade' message. We shall assume that $\phi^*_{S_4}$ is incomplete (i.e., an active cascade message). In Section 6 we shall elaborate on the issues of message types and message scheduling.

# 6   MESSAGE SCHEDULING

In a junction tree involving only exact computations and active messages, message scheduling can be designed as in Hugin, following a strict sequential one-branch-at-a-time schedule in both the inward an the outward pass. The possible need for cascading makes such a rigid scheme inadequate for hybrid junction trees.

Instead we shall take a more localized, object-oriented approach, where each object (universe) performs book-keeping regarding incoming and outgoing active messages. Thus, for example, if a universe has received active messages from all of its neighbours except one, it can send an active message to that neighbour. When each universe has sent and received active messages to/from all of its neighbours, message passing stops and equilibrium has been established.

When sending a cascade message it should obviously be checked if the message is active, and if so, appropriate book-keeping should take place. For example, the cascade message from $GIBBS_2$ to $DE_1$ is active, and once this message has been absorbed by $DE_1$, the cascade message from $DE_1$ to $GIBBS_1$ is also active, whereas the subsequent cascade messages from $GIBBS_1$ are non-active; but they may nevertheless be required, since $GIBBS_1$ and $GIBBS_3$ share the variables PB1, D/A2, and PA2.

In summary, what we need to know to build a message-scheduling mechanism for HUGS is the following.

(1) A message can be one of three kinds:

   (a) active non-cascade,
   (b) non-active cascade, or
   (c) active cascade.

(2) When an active message is received, the recipient notes that an active message has been received from the sender.

(3) When an active message is sent, the sender notes that an active message has been sent to the recipient.

(4) A universe may send an active message to a neighbour when and only when (i) it has received active messages from all of its neighbours except possibly from the recipient, and (ii) no cascading is taking place.

(5) When a universe receives a cascade message, it must absorb the message, and start sampling if the universe is of type GIBBS.

(6) When a GIBBS universe has completed sampling or a DE universe has absorbed a cascade message, the universe must send cascade messages to all of its neighbours

   (a) from which it has not yet received messages,
   (b) in the subtrees of which there are GIBBS universes, and
   (c) for which incomplete messages will be generated with support smaller than the support of the corresponding marginal of the recipient.

Note that obedience to the second premise of Rule (4) is easily observed as long as the message passing is implemented on a sequential computer. A parallel implementation requires some sort of central control mechanism to prevent universes from sending (including generating) active, non-cascade messages whenever cascading takes place.

# 7   FINISHING CASCADING

The cascading initiated by $GIBBS_2$ must be completed before any other message-sending activities take place.

First, the (active cascade) message (1) is absorbed by $DE_1$, assuming that evidence $\mathcal{E}_{LA1} = (0, 1, 0)$ is entered into $DE_1$:

$$\phi'_{DE_1}(x) = \phi_{DE_1}(x) * \mathcal{E}_{LA1}(x_{LA1}) * \phi^*_{S_4}(x_{S_4}). \quad (2)$$

Assume now that Rule (6) in Section 6 applies. That is, $DE_1$ must send a cascade message to $GIBBS_1$. Note that this message is active.

$GIBBS_1$ absorbs this message, and according to Rules (5)–(6), $GIBBS_1$ must perform sampling and send cascade messages to those of its neighbours, $DE_2$ and $DE_3$, for which Rule (6) applies. Gibbs sampling is first conducted to turn $GIBBS_1$ into a DE universe.

As described in Section 4.1, we first need to find a legal initial state, $x^0_{GIBBS_1}$, fulfilling the requirement $\phi_{GIBBS_1}(x^0_{GIBBS_1}) > 0$. We might start by drawing an $x^0_{S_1}$ from $\phi^*_{S_1}$. Thus $x^0_{GIBBS_1} = (x^0_{S_1}, x^0_{GIBBS_1 \setminus S_1})$, where $x^0_{GIBBS_1 \setminus S_1}$ could be determined as described in Section 4.1 with $X_{S_1}$ clamped to $x^0_{S_1}$. If no legal $x^0_{GIBBS_1 \setminus S_1}$ exist, we could draw another $x^0_{S_1}$ from $\phi^*_{S_1}$, etc.

Using the subgraph induced by the variables of $GIBBS_1$ (cf. Figure 1), Table 1, and sampling order PB1, D/A2,

PA2, $S_1 = \{\text{LA1}, \text{PA1}, \text{B}\}$ we generate the $i$'th sample as follows

(1) Draw $x_{\text{PB1}}^i$ from

$$\phi(X_{\text{PB1}} \mid x_{\text{B}}^{i-1}, x_{\text{D/A2}}^{i-1}) \propto$$
$$\phi(X_{\text{PB1}} \mid x_{\text{B}}^{i-1}) * \phi(x_{\text{D/A2}}^{i-1} \mid x_{\text{PA1}}^{i-1}, X_{\text{PB1}}).$$

(2) Draw $x_{\text{D/A2}}^i$ from

$$\phi(X_{\text{D/A2}} \mid x_{\text{PA1}}^{i-1}, x_{\text{PB1}}^i, x_{\text{PA2}}^{i-1}) \propto$$
$$\phi(X_{\text{D/A2}} \mid x_{\text{PA1}}^{i-1}, x_{\text{PB1}}^i) *$$
$$\phi(x_{\text{PA2}}^{i-1} \mid x_{\text{LA1}}^{i-1}, x_{\text{PA1}}^{i-1}, X_{\text{D/A2}}).$$

(3) Draw $x_{\text{PA2}}^i$ from

$$\phi(X_{\text{PA2}} \mid x_{\text{LA1}}^{i-1}, x_{\text{PA1}}^{i-1}, x_{\text{D/A2}}^i).$$

(4) Draw $x_{S_1}^i = (x_{\text{LA1}}^i, x_{\text{PA1}}^i, x_{\text{B}}^i)$ from

$$\phi_{S_1}^*(X_{\text{B}}, X_{\text{LA1}}, X_{\text{PA1}}) * \phi(x_{\text{PB1}}^i \mid X_{\text{B}}) *$$
$$\phi(x_{\text{D/A2}}^i \mid X_{\text{PA1}}, x_{\text{PB1}}^i) *$$
$$\phi(x_{\text{PA2}}^i \mid X_{\text{LA1}}, X_{\text{PA1}}, x_{\text{D/A2}}^i).$$

When sampling has been completed, we are left with a list of possible configurations $\mathcal{X}_{\text{GIBBS}_1}^* \subseteq \mathcal{X}_{\text{GIBBS}_1}$ and a list of associated weights.

Assume now that no cascading to DE$_2$ is required and that a cascade message, $\phi_{S_3}^*$, must be sent to DE$_3$. Note that this message is not active; following the terminology of Section 6, it is a non-active cascade message.

The absorption of $\phi_{S_3}^*$ into DE$_3$, producing $\phi_{\text{DE}_3}^*$, is similar to the absorption of $\phi_{S_4}^*$ into DE$_1$ (cf. (2)). We assume that a non-active cascade message, $\phi_{S_5}^* = \sum_{\text{DE}_3 \backslash S_5} \phi_{\text{DE}_3}^*$, must be sent to GIBBS$_3$. The absorption of $\phi_{S_5}^*$ into GIBBS$_3$ and subsequent generation of $N$ samples is similar to the message absorption and sampling that took place in GIBBS$_1$, except that evidence $\mathcal{E}_{\text{D/A3}} = (0, 1)$ is taken into account through clamping D/A3 to the value 'no'.

Now, no further cascading is required, since all three GIBBS universes have been transformed into DE universes at this point.

## 8    FINISHING MESSAGE PASSING

Now we complete the inward pass: the only universes in position to send active messages are the leaf universes DE$_2$, DE$_4$, and DE$_5$. Assume that they send messages in that order. Next, DE$_6$ absorbs the evidence $\mathcal{E}_{\text{O}} = (1, 0)$ and sends an active message to (the

DE universe) GIBBS$_3$, which in turn sends one to DE$_3$. Finally, DE$_3$ sends an active message to GIBBS$_1$, finishing the inward pass. Figure 3 illustrates the inward process, where solid arrows indicate active messages, dashed arrows indicate non-active cascade messages, and the labels attached to the arrows indicate the order in which the messages were sent.

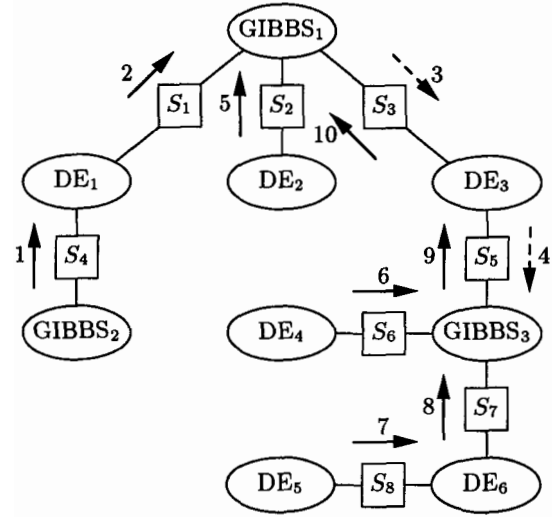

Figure 3: Active cascade messages have been sent from GIBBS$_2$ to DE$_1$ (1) and from DE$_1$ to GIBBS$_1$ (2). Non-active cascade messages have been sent from GIBBS$_1$ to DE$_3$ (3) and from DE$_3$ to GIBBS$_3$ (4) (dashed arrows). Remaining messages, (5)–(10), are all active non-cascade messages. To complete the propagation (i.e., to perform the outward pass), GIBBS$_1$ must send active messages to its neighbours, which in turn must send to their neighbours, etc.

Having finished inward message passing in the hybrid junction tree, all universes will be of type DE; thus, outward message passing in a hybrid junction tree is performed as in conventional DE-type trees.

## 9    DISCUSSION

For illustration purposes the message scheduling in our example made a GIBBS universe send the first message. In general, however, this will probably be suboptimal, since, as mentioned above, the more information available before sampling takes place the better. Thus, an optimal message scheduling strategy will probably select DE universes whenever possible. Thus, in our example a better schedule would probably let DE$_2$, DE$_4$, and DE$_6$ send messages before either GIBBS$_2$ or GIBBS$_3$ send the next message.

As an indication of the anticipated ratio between the numbers of GIBBS and DE universes, we have investigated a subnetwork of the MUNIN network (Andreassen, Jensen, Andersen, Falck, Kjærulff, Woldbye, Sørensen, Rosenfalck & Jensen 1989) consisting of 1041 nodes and 876 universes in a corresponding junction tree. Letting all universes with tables containing more than $100,000$ entries be GIBBS universes, we find that 41 of the 876 universes should be of type GIBBS. The tables of these universes (constituting less than 5% of the universes) take up 91.7% of the total storage engaged by universe tables. Assuming that we generate 10,000 samples in each of the 41 GIBBS universes, we still face a reduction of storage requirement by about 90%. In absolute values, this means a reduction from 66 Mbytes to 7 Mbytes, using 32-bit representation of floating point numbers.

Experience shows that the ratio between the number of 'very large' universes and the number of 'smaller' universes in the above example is typical. That is, a small number of universes take up the majority of the total storage engaged. Thus, to obtain a large reduction of storage requirement, only a small proportion of the universes need to be of type GIBBS.

Supposedly, the present paper represents the very first attempt to combine exact and approximate inference in Bayesian networks. So, obviously, a large number of theoretical as well as practical issues need to be addressed. On the practical side, a thorough comparison with standard Gibbs and exact methods, will be conducted in the very near future.

Regarding theoretical work, a large number of issues can be addressed. A few important ones are development of methods providing variance estimates of the posteriors, development of methods to handle various kinds of continuous distributions, and development of alternative sampling schemes.

## Acknowledgements

I am imdebted to my co-authors, A. Philip Dawid and Steffen L. Lauritzen, for providing the foundation of the present paper in the paper of Dawid et al. (1994). Further, I grateful to Steffen L. Lauritzen and anonymous referees for providing valuable comments on earlier drafts of the paper. Finally, I thank the remaining members of the ODIN group at Aalborg University (http://www.iesd.auc.dk/odin) for providing a stimulating environment. This research was supported by the Danish Research Councils through the PIFT programme.

## References

Andersen, S. K., Olesen, K. G., Jensen, F. V. & Jensen, F. (1989). HUGIN — A shell for building Bayesian belief universes for expert systems, *Proceedings of the Eleventh International Joint Conference on Artificial Intelligence*, pp. 1080–1085.

Andreassen, S., Jensen, F. V., Andersen, S. K., Falck, B., Kjærulff, U., Woldbye, M., Sørensen, A. R., Rosenfalck, A. & Jensen, F. (1989). MUNIN — an expert EMG assistant, *in* J. E. Desmedt (ed.), *Computer-Aided Electromyography and Expert Systems*, Elsevier Science Publishers B. V. (North-Holland), Amsterdam, chapter 21.

Dawid, A. P. (1992). Applications of a general propagation algorithm for probabilistic expert systems, *Statistics and Computing* **2**: 25–36.

Dawid, A. P., Kjærulff, U. & Lauritzen, S. L. (1994). Hybrid propagation in junction trees, *Proceedings of the Fifth International Conference on Information Processing and Management of Uncertainty in Knowledge-Based Systems (IPMU)*, Cité Internationale Universitaire, Paris, pp. 965–971.

Geman, S. & Geman, D. (1984). Stochastic relaxation, Gibbs distributions, and the Bayesian restoration of images, *IEEE Transactions on Pattern Analysis and Machine Intelligence* **6**(6): 721–741.

Jensen, C. S., Kong, A. & Kjærulff, U. (1995). Blocking-Gibbs sampling in very large probabilistic expert systems, *International Journal of Human-Computer Studies* . Special Issue on Real-World Applications of Uncertain Reasoning. To appear.

Jensen, F. V. (1988). Junction trees and decomposable hypergraphs, *Research report*, Judex Datasystemer A/S, Aalborg, Denmark.

Jensen, F. V., Lauritzen, S. L. & Olesen, K. G. (1990). Bayesian updating in causal probabilistic networks by local computations, *Computational Statistics Quarterly* **4**: 269–282.

Lauritzen, S. L. & Spiegelhalter, D. J. (1988). Local computations with probabilities on graphical structures and their application to expert systems, *Journal of the Royal Statistical Society, Series B* **50**(2): 157–224.

Thomas, A., Spiegelhalter, D. J. & Gilks, W. R. (1992). BUGS: A program to perform Bayesian inference using Gibbs sampling, *in* J. M. Bernardo, J. O. Berger, A. P. Dawid & A. F. M. Smith (eds), *Bayesian Statistics 4*, Oxford University Press, Oxford, UK, pp. 837–842.